\documentclass[letterpaper, 10 pt, conference]{ieeeconf}  

\IEEEoverridecommandlockouts                              

\usepackage{graphicx}
\usepackage{amsmath} 
\usepackage{amssymb}  
\usepackage{hyperref}

\usepackage{graphicx}
\usepackage{hyperref}
\usepackage{cleveref}
\usepackage{keyval}
\usepackage[ruled,vlined]{algorithm2e}
\usepackage{amsmath}
\usepackage{amssymb}
\usepackage{color}

\newcommand{\idest}{{\it i.e.}, }
\newcommand{\exempli}{{\it e.g.}, }

\title{\LARGE \bf
A Generative Approach for Socially Compliant Navigation\thanks{This material is based upon work supported in part through collaborative participation in the Robotics Consortium sponsored by the U.S Army Research Laboratory under the Collaborative Technology Alliance Program, Cooperative Agreement W911NF-10-2-0016. This work should not be interpreted as representing the official policies, either expressed or implied, of the Army Research Laboratory of the U.S. Government. The U.S. Government is authorized to reproduce and distribute reprints for Government purposes notwithstanding any copyright notation herein.}
}

\author{Chieh-En Tsai and Jean Oh$^{1}$\thanks{$^{1}$The authors are with the Robotics Institute at Carnegie Mellon University.}
}

\begin{document}

\maketitle
\thispagestyle{empty}
\pagestyle{empty}

\begin{abstract}

Robots navigating in human crowds need to optimize their paths not only for 
their task performance but also for their compliance to social norms. One of the key challenges in this context is the lack of standard metrics for evaluating and optimizing a socially compliant behavior. 
Existing works in social navigation can be grouped according to the differences in their optimization objectives. 
For instance, the reinforcement learning approaches tend to optimize on the \textit{comfort} aspect of the socially compliant navigation, whereas the inverse reinforcement learning approaches are designed to achieve \textit{natural} behavior.
In this paper, we propose NaviGAN, a generative navigation algorithm that jointly optimizes both of the \textit{comfort} and \textit{naturalness} aspects. Our approach is designed as an adversarial training framework that can learn to generate a navigation path that is both optimized for achieving a goal and for complying with latent social rules. 
A set of experiments has been carried out on multiple datasets to demonstrate the strengths of the proposed approach quantitatively. We also perform extensive experiments using a physical robot in a real-world environment to qualitatively evaluate the trained social navigation behavior. The video recordings of the robot experiments can be found in the link: https://youtu.be/61blDymjCpw.

\end{abstract}

\section{Introduction}
\begin{figure}[t]
    \centering%
    \includegraphics[scale=0.15]{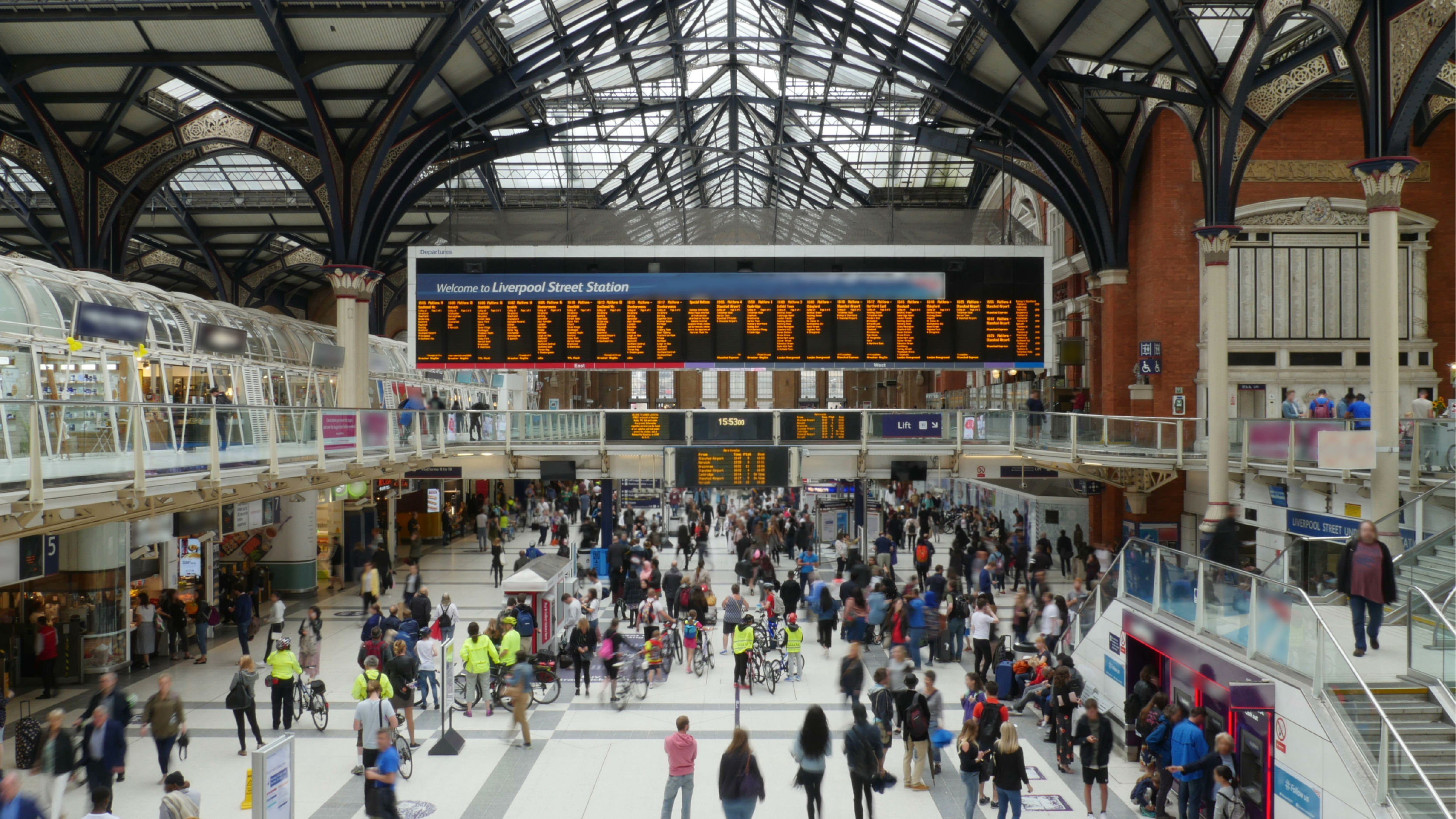}%
    \caption{An example showing the complexity of navigating in human crowds.} 
    \label{fig:intro}
    \vspace{-15pt}
\end{figure}

In the context where robots co-exist and collaborate with people, social compliance becomes an important objective in designing robot behaviors. 
socially compliant robots would 
consider not only which actions will lead to a successful completion of their tasks but also whether those actions will be acceptable within a social context.
For instance, when navigating in a crowd as illustrated in Fig.~\ref{fig:intro}, a socially compliant robot must have an ability to perceive and understand crowd behavior to plan its future trajectory to reach its destination while maintaining an appropriate distance from other pedestrians.

We note	that our target	problem	is different from pedestrian prediction~\cite{alahi2016social,vemula2017social,gupta2018social} where the objective does not include a goal. 
Instead, we aim to develop a navigation algorithm where an agent needs to plan its trajectory to reach a specific goal (or destination) when it is situated in a crowd.

There are three areas of research for improving user acceptance in social navigation~\cite{KRUSE20131726}: 
\textit{ (i) Comfort} is the absence of annoyance and stress for humans in interaction with robots; \textit{ (ii) Naturalness}, the similarity between robots and humans in low-level behavior patterns; and \textit{ (iii) Sociability}, the adherence to explicit high-level cultural conventions. In this paper, we discuss the first two terms that are most relevant to navigation, and leave the third for future work.

Two main streams of research in socially compliant navigation are reinforcement learning (RL) methods~\cite{everett2018motion, long2018towards, chen2017decentralized, chen2017socially, chen2018crowd} and inverse reinforcement learning (IRL) methods~\cite{kretzschmar2016socially, vasquez2014inverse, okal2016learning, kitani2012activity, pfeiffer2016predicting}.
A major drawback of the RL approaches comes from the fact that it is nontrivial to elicit a sophisticated reward function that will lead to a satisfying policy for both comfort and naturalness. Besides, it is hard to roll out episodes for training. For instance, exploring in a real-world environment is expensive and can be unsafe for both robots and human pedestrians; however, training using a simulator generally faces the domain-mismatch issues because none of the existing simulators reflect human behaviors perfectly.
By contrast, the IRL approaches attempt to learn from human demonstrations. In IRL, the lack of negative examples, such as collisions, can lead to a model that only focuses on naturalness without considering comfort or safety. In addition, a careful feature engineering is generally necessary to achieve reasonable performance as in~\cite{kretzschmar2016socially}.  

In this work, we present NaviGAN--a generative approach for performing socially compliant robotic navigation that considers both \textit{comfort} and \textit{naturalness} of social compliance.
Following the social force model~\cite{helbing1995social} where an agent's decision consists of three types of forces, we separately model each force in our approach; that is, an agent's \textit{intention} force that directs to its destination, \textit{social interaction} force that is governed by social norms, and \textit{fluctuation} force that can arise from natural randomness. 
We design a separate encoder-decoder~\cite{sutskever2014sequence} model using a Long Short-Term Memory (LSTM)\cite{hochreiter1997long} to represent each different type of influence (or force) on navigation decisions.
This separation enhances the algorithm's interpretability by being able to show the effects of each type of force. 
At the same time, LSTM-based models generally suffer from data bias resulting in a model that can unfairly penalize equally-good but less frequently observed behaviors. To address this issue, motivated by 
generative adversarial networks (GANs)~\cite{goodfellow2014generative} based method, we pair the force generators with a social-aware discriminator that learns to identify generic, socially compliant navigation paths.


To evaluate the performance of NaviGAN,
we deploy our algorithm on a mobile robot to perform extensive real-world experiments including both qualitative assessments and the A/B testing. Additionally, we report on the feedback from the  pedestrians who participated in the experiments. 

The main contributions of this paper are: 1) we propose a socially compliant navigation algorithm covering both comfort and naturalness; 2) the proposed approach is purely data-driven and does not need extensive feature engineering as seen in existing works; 3) the proposed approach is highly interpretable; and 4) we carry out a set of real robot experiments and share the findings. 

\section{Related Works}
In current literature, the majority of the reinforcement learning (RL)~\cite{kober2013reinforcement, szepesvari2010algorithms, kaelbling1996reinforcement,sutton1998introduction} based methods~\cite{everett2018motion, long2018towards, chen2017decentralized, chen2017socially, chen2018crowd} for socially compliant navigation are designed to optimize the comfort aspect. Chen et. al~\cite{chen2017socially, chen2018crowd} propose to use Deep V-learning in combination with a hand-crafted reward function that penalizes collisions to learn the value network that induces an optimal policy that maximizes the expected reward. The reward function to evaluate and optimize the social awareness of a robot policy used by~\cite{chen2017decentralized, chen2017socially, chen2018crowd} is defined as Eq.~\ref{eq:reward} %
where $R_t$ is the reward at time $t$; and $d_t$, the minimum separation distance between a robot and a human during the execution of the last action. Such a reward function captures the dynamic collision avoidance aspect that only satisfies the safety requirement--or the minimum level of the comfort requirement--of socially compliant navigation. As socially compliant navigation requires a richer model than only penalizing on collisions and uncomfortable distances, the optimal social reward $R_t$ is non-trivial to specify.

\vspace{-5pt}
\begin{equation}
    R_t = \begin{cases}
    -0.25 &, \text{if } d_t \leq 0 \text{ meter}\\
    -0.1+\frac{d_t}{2} &, \text{else if } d_t < 0.2 \text{ meter}\\
    1 &, \text{else if reach goal} \\
    0 &, \text{otherwise}
    \end{cases}
    \label{eq:reward}
\end{equation}

To cover the naturalness aspect, another stream of research takes the ``learning from demonstration'' approach.  Specifically, ~\cite{kretzschmar2016socially, vasquez2014inverse, okal2016learning, kitani2012activity, pfeiffer2016predicting} utilizes inverse reinforcement learning (IRL)~\cite{abbeel2011inverse, abbeel2004apprenticeship, ziebart2008maximum, ng2000algorithms} to jointly learn a reward function and a policy for socially compliant navigation. They take a purely data-driven approach to uncover the latent aspects of social navigation from expert demonstrations. As a policy is learned directly from human demonstrations, this approach can capture human navigation patterns, which makes it suitable for training the naturalness of socially compliant navigation. However, as the expert demonstrations tend to lack negative examples--\exempli pedestrians normally rarely collide--the model can only implicitly learn the safety aspect without given any explicit guidance.
Such a low awareness of safety makes the IRL-based methods hard to generalize as the number of pedestrians increases. Due to this reason, most of the previous works in IRL for socially compliant navigation only learn from and test on a small number of pedestrians.

In order to plan a long-term human-aware path to support challenging robotics scenarios, \exempli robot-guides or self-driving delivery services, the  social-aware path prediction problem recently gains interests among researchers. Social-aware path prediction refers to the ability to predict a trajectory that each social agent will take given their previous trajectories as observation. In order to correctly predict the trajectory for each social agent, we need to model the interactive behavior among social agents including collaborative behavior. Existing works~\cite{antonini2006discrete, luber2010people, alahi2014socially, huang2014action, alahi2016social,vemula2017social,gupta2018social} in this area focus on 
modeling local, reactive behavior of human pedestrians as opposed to planning for navigation among crowds.

Toward the goal of developing a navigation algorithm that can produce socially compliant paths, our approach, NaviGAN, differs from existing works as follows: 
%
First, rather than mimicking human demonstrations as in pedestrian trajectory prediction works, \exempli using the L2 loss, we propose an objective function that is geared to learn generalized behavior from demonstrations. 
Next, although deep networks~\cite{alahi2016social,vemula2017social,gupta2018social} are the winning approaches here, their lack of interpretability is a common issue as in other domains. Finally, the L2 metric for evaluation captures only the difference of generated path from ground truth recording. As the concept of socially compliance is complicated and the ground truth is not the only solution that is socially compliant, we seek more metrics for evaluating the performance. 

\section{Approach}
In this paper, a \textit{target agent} refers to the agent that performs a navigation task in a populated environment. We assume that a target agent knows its goal (or destination) and is able to detect and track pedestrians.

\subsection{Formulation}
Let $\mathcal{D}$ denote a set of agents (pedestrians) that are involved in the time sequence $\{1, \dots, T_{obs}, \dots, T_{end}\}$. Given target agent $i$ in $\mathcal{D}$, let $g_i$ denote the goal states of agent $i$; $X_i = \{x_i^1, \dots, x_i^{T_{obs}}\}$, the past state sequence of agent $i$ for time steps $t=1 \dots T_{obs}$; and $Y_i = \{y_i^{T_{obs}+1}, \dots, y_i^{T_{end}}\}$, the demonstrated navigation state sequence provided by agent $i$, where $x_i^t$ and $y_{i}^{t'}$ denote the state of agent $i$ at time step $t$ and $t'$, respectively. For each agent $i$ in $\mathcal{D}$, given a sequence of past trajectories of all agents $\{X_j: j \in \mathcal{D}\}$ and the target agent's goal state $g_i$, we aim to generate a navigation state sequence $\hat{Y}_i = \{\hat{y}_i^{T_{obs}+1}, \dots, \hat{y}_i^{T_{end}}\}$ up to time step $T_{end}$ with an objective of navigating toward the agent's goal state $g_{i}$ in a socially compliant manner. 

\subsection{Overview}

Following the theory of social force \cite{helbing1995social}, we propose NaviGAN, a generative adversarial network architecture for social navigation, that represents an agent's mixed intention to reach a goal while trying to comply with social norm of crowd navigation behavior. 

\begin{figure*}[t]
\begin{center}%
\vspace{2pt}
\includegraphics[scale=0.40]{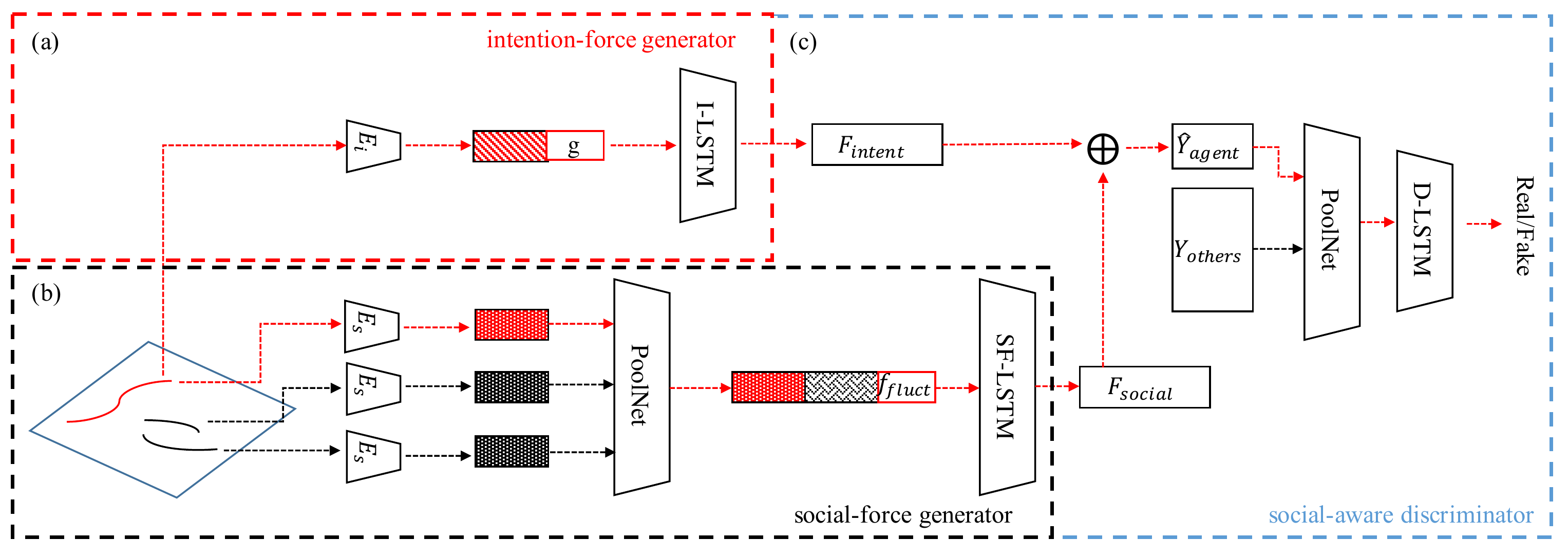}%
\vspace{-10pt}
\end{center}%
    \vspace{-4pt}
   \caption{Overview of the proposed model. NaviGAN is composed of three major building blocks: 1. Intention-force generator (block (a), section \ref{intention}) that models the intention force $\Vec{F}_{intent}$ toward goal state; 2. Social-force generator (block (b), section \ref{social}) that models the social force $\Vec{F}_{social}$ and fluctuation force $\Vec{F}_{fluct}$; and 3. Social-aware discriminator (block (c), section \ref{discriminator}) that discovers the latent social aspects from discriminating between expert demonstrations and navigation behavior generated by the generator. The red color indicates that the vector holds data relevant to the target agent, and the black color is used for other agents.}%
   \vspace{-17pt}
\label{fig:overview}%
\end{figure*}

An overview of our proposed model is shown in Fig. \ref{fig:overview}. NaviGAN consists of three building blocks: 
(i) Intention-force $\Vec{F}_{intent}$ generator 
that models an agent's intention for reaching a destination, 
(ii) Social-force generator 
that models the social force $\Vec{F}_{social}$ and fluctuation force $\Vec{F}_{fluct}$, and 
(iii) Social-aware discriminator 
that discover the latent social influences from discriminating those navigation paths generated by the generator against expert demonstrations.
Each building block is described in detail in section \ref{intention} to section \ref{discriminator}.


\subsection{Intention-Force Generator} \label{intention}
The purpose of the intention-force generator is to generate a path (or a sequence of states) $\Tilde{Y_i} = \{\Tilde{y_i}^{T_{obs}+1}, \dots, \Tilde{y_i}^{T_{end}}\}$ toward goal state $g_{i}$, conditioned on $g_i$ and the past state sequence $X_i$ of the target agent. 
Note that this part represents self-interested decision making where the benefits of other agents are not considered. 
As shown in block (a) of Fig. \ref{fig:overview}, the intention-force generator follows an encoder-decoder design. We encode the past state sequence $X_i$ of target agent $i$ using an encoder LSTM $E_i$. The initial hidden state $h^0_i$ and cell state $c^0_i$ of the target agent are set with a zero vector. For time steps $t = 1 \dots T_{obs}$, we perform the following update operation to encode the input sequence:
\begin{equation}
    [h^t_i; c^t_i] = E_i(x_i^t, [h^{t-1}_i; c^{t-1}_i]), \label{eq:Encode}
\end{equation}
where the operation $[u; v]$ denotes concatenating vectors $u$ and $v$. We then inject goal state $g_i$ to the resulting hidden state to initialize the hidden state $\mathrm{h}^{T_{obs}}_i$ for the decoder, known here as Intention-LSTM (I-LSTM): $\mathrm{h}^{T_{obs}}_i = [h^{T_{obs}}_i; g_i]$.
For timesteps $t = T_{obs}+1 \dots T_{end}$, the decoding process can be described by:
\begin{align*}
    [\mathrm{h}^t_i; \mathrm{c}^t_i] & = \text{I-LSTM}(\Tilde{y}^{t-1}_i, [\mathrm{h}^{t-1}_i; \mathrm{c}^{t-1}_i]) \\
    \Tilde{y}^t_i & = \text{NN}_{spatial}(\mathrm{h}^t_i),
\end{align*}
where $\Tilde{y}^{T_{obs}}_i$ is initialized with $x_i^{T_{obs}}$. $\text{NN}_{spatial}$ is a multilayer perceptron to map the hidden state of I-LSTM $\mathrm{h}^t_i$ to state prediction output $\Tilde{y}^t_i$. 

\subsection{Social-Force Generator} \label{social}
Fig. \ref{fig:overview} illustrates the social-force generator.
 
Given all of the observed state sequences $X_\mathcal{D} = \{X_j: j \in \mathcal{D}\}$ in demonstration $\mathcal{D}$, the social-force generator will generate a sequence of social forces $\{\Vec{f}_i^{T_{obs}+1}, \dots, \Vec{f}_i^{T_{end}}\}$. The composite social force $\Vec{f}_i^t$ is the ensemble of social force and fluctuation force that models the social responses of the target agent to all of other agents in the scene as well as the stochastic nature of human navigation policies. 
The social-force generator is an encoder-pool-decoder design. 
It performs encoding for all of the agents in the scene and then utilizes a special pooling mechanism before decoding. The idea of designing the social-force generator is to predict the social interactions between agents purely based on their relative positions and previous state sequences. We encode all of the observed state sequences $X_j \in X_\mathcal{D}$ using an encoder LSTM $E_s$. For simplicity, we reuse the notations of $h$ and $c$ to represent the hidden state and cell state related to $E_s$; $\mathrm{h}$ and $\mathrm{c}$ to represent the hidden state and cell state of SocialForce-LSTM (SF-LSTM). The initial hidden state $h^0_j$ and cell state $c^0_j$ of agent $j$ are initialized with a zero vector. Using the same encoding process as Eq. \ref{eq:Encode}, we obtain the encoding vector $h_j^{T_{obs}}$ of each $X_j$.

For the efficiency and simplicity, we adopt a similar pooling mechanism as \cite{gupta2018social}. The \textit{PoolNet} module will consider the displacement--\idest the vector from target agent's coordinate to other agent's coordinate --from the target agent to all of other agents when generating a pooled context vector that is composed of encoding vectors of all other agents and the target agent. Let $d_{ij}^{T_{obs}}$ denote the displacement from target agent $i$ to agent $j$ at time $T_{obs}$. We obtain a displacement-sensitive context embedding $e_{ij}^{T_{obs}}$ for each $X_j$ by: 1) feeding displacement $d_{ij}^{T_{obs}}$ through a position encoding neural network $\text{NN}_{pos}$, and then 2) concatenating the output with $h_j^{T_{obs}}$ and feeding through the displacement-sensitive embedding network $\text{NN}_{embed}$. The process is formally described in the following equation:
\begin{equation*}
    e_{ij}^{T_{obs}} = \text{NN}_{embed}([h_j^{T_{obs}}; \text{NN}_{pos}(d_{ij}^{T_{obs}})]).
\end{equation*}
After we obtain the displacement-sensitive context embedding $e_{ij}^{T_{obs}}$ for each agent, we aggregate them into a single vector $\mathrm{V}^{T_{obs}}_i$ using max pooling as defined as follows: 
\begin{equation}
    \mathrm{V}^{T_{obs}}_i [idx] = \max_{j\in \mathcal{D} \setminus \{i\}} e_{ij}^{T_{obs}} [idx], \label{eq:maxpool}
\end{equation}
where $v[idx]$ denotes the operation that indexes the $idx^{th}$ entry of vector $v$. The final vector $\mathrm{V}^{T_{obs}}_i$ is the ensemble of all of the information from other agents that leads the social-force generator to determine the target agent's social responses to other agents.

\subsection{Fluctuation Force Generator} \label{ff}
Finally, we initialize the hidden state $\mathrm{h}^{T_{obs}}_i$ of the SF-LSTM by concatenating the target agent's context vector $h_i^{T_{obs}}$, the aggregated context vector $\mathrm{V}^{T_{obs}}_i$, and a random noise vector $f_{fluct}$ sampled from $\mathcal{N}(0, 1)$ which will serve as the random seed for modeling the stochastic $\Vec{F}_{fluct}$. 
\begin{equation*}
    \mathrm{h}^{T_{obs}}_i = [h_i^{T_{obs}}; \mathrm{V}^{T_{obs}}_i; f_{fluct}]
\end{equation*}
The cell state $\mathrm{c}^{T_{obs}}_i$ is initialized with a zero vector. For time steps $t = T_{obs}+1 \dots T_{end}$, the decoding process can then be described by the following equations:
\begin{align*}
    [\mathrm{h}^t_i; \mathrm{c}^t_i] & = \text{SF-LSTM}(\Vec{f}_i^{t-1}, [\mathrm{h}^{t-1}_i; \mathrm{c}^{t-1}_i]) \\
    \Vec{f}_i^{t} & = \text{NN}_{social}(\mathrm{h}^t_i),
\end{align*}
where $\Vec{f}_i^{T_{obs}}$ is initialized with a zero vector and $\text{NN}_{social}$ is a multilayer perceptron to map the hidden state of SF-LSTM $\mathrm{h}^t_i$ to social force prediction $\Vec{f}_i^{t}$. 

To simply the implementation, the fluctuation force is concatenated at the end of the social force generating process as shown in Fig.~\ref{fig:overview}.
%
\subsection{Social-Aware Discriminator} \label{discriminator}
The way-point $\hat{y}_i^t$ that the target agent should reach by executing an action at time $t$ is decided by combining the intention force and social forces: $\hat{y}_i^t = \Tilde{y}_i^t + \Vec{f}_i^t$.

Based on the intuition that there can be multiple optimal behaviors that may not be present in expert demonstrations, we take the adversarial training approach as opposed to mimicking demonstrations by optimizing on the negative likelihood of demonstrations. 
The prediction process of social-aware discriminator is illustrated in block (c) of Fig. \ref{fig:overview}. Assume the input target agent state sequence is $\{s_i^t: t=1\dots T_{end}\}$, and other agent state sequence is $\{s_j^t: t=1\dots T_{end}\}$. At each time step $t$ we aggregate the target agent state $s_i^t$ with other agents' state $s_j^t$ into one target-centric, displacement-sensitive context embedding $v_i^t$ using the PoolNet module. The sequence of target-centric embedding $\{v_i^t: t=1\dots T_{end}\}$ is then fed to Discriminator-LSTM (D-LSTM) to encode the sequence into one unified vector $h_i^{T_{end}}$ which is the last hidden state of D-LSTM. Base on $h_i^{T_{end}}$, the discriminative score is predicted to identify whether the input sequence is generated by expert demonstration or generators.


\subsection{Enforce Comfort}
The adversarial training framework discovers latent social aspects to guide generators into producing policy that mimics the low-level behavior patterns demonstrated by the recording. Though this covers the \textit{naturalness} aspect of socially compliant navigation, the \textit{comfort} aspect is not explicitly governed. Therefore, we include resistance loss $\mathcal{L}_{resist}$ that explicitly penalizes behaviors where predicted trajectory comes closer to other pedestrian than a safety distance threshold.
\begin{equation}
    \mathcal{L}_{resist} = \lVert \max (d_{safe}-d_o, 0) \rVert_2
\end{equation}

$d_{safe}$ is the safety distance threshold, and $d_o$ is the distance from target agent to other agent. By including $\mathcal{L}_{resist}$ in our training framework, we jointly optimize both \textit{comfort} and \textit{naturalness}.

\section{Experiments}

\begin{figure}

    \centering%
    \includegraphics[scale=0.3]{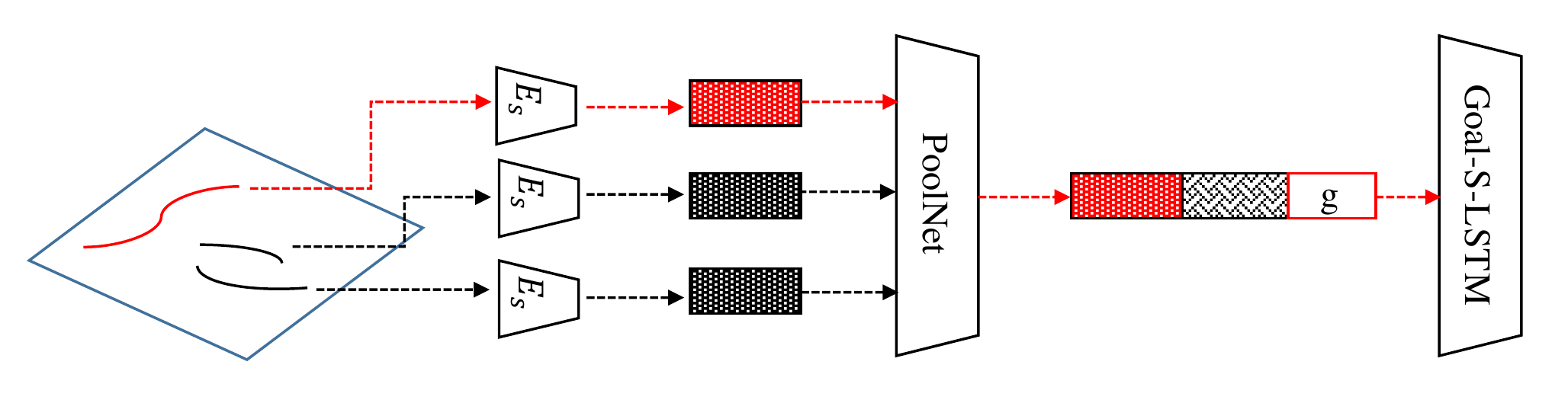}
    \caption{The architecture of Goal-Social-LSTM. Goal state information $g$ is directly injected after PoolNet layer to predict state sequence condition on goal state.}%
    
    \label{fig:goal_social}%
    \vspace{-15pt}
\end{figure}

For the consistency throughout the experiments, we set the dimension of hidden states of all of the encoder and decoder LSTMs to 32. We iteratively train the generators and discriminator with batch size 32, i.e., the number of target agent for each batch, for 500 epochs using the Adam \cite{kingma2014adam} optimizer with learning rate of 0.001.

\subsection{Experimental setup}
\textbf{Datasets:} We conduct our experiments on two publicly available datasets: ETH \cite{pellegrini2010improving} and UCY \cite{leal2014learning}. In total, there are 5 sets of data (ETH, HOTEL, UNIV, ZARA1, ZARA2) with 4 different scenes and 1,536 pedestrians in crowded settings. Each pedestrian is associated with a unique index, annotated at 2.5 fps. We set the $T_{obs} = 8$ (3.2 seconds) and $T_{pred} = T_{end} - T_{obs} = 12$ (4.8 seconds). To encourage a model to learn long-term, complex social behavior, we follow the common practice \cite{alahi2016social,gupta2018social} of selectively choosing the pedestrians that stay in the receptive field for longer than $T_{pred}$ to be our target agents, and learn the policy through their demonstration. 

\textbf{Coordinate frames:} As we model the interactions between a target agent and other agents, we use a target-centric coordinate frame where its origin is on the point $x^0_i$ in world coordinates. 
In comparison to world coordinates, the use of target-centric coordinates tends to result in a more general model by preventing the model from learning location-specific biases.

\textbf{Setup for long-term navigation tests:} 
To test long-term navigation, we set the goal state to be the target agent's position $3\cdot T_{pred}$ (14.4 seconds) into the future and roll out the episode with a cut-off horizon at $5\cdot T_{pred}$ (24 seconds). During execution, whenever a target agent arrives within 0.5m from its goal state, we end the episode and mark the navigation as a success. At each time step, we use a target agent's previous predictions and others' observed trajectories as input to predict the next action and execute in order to simulate a realistic navigation.

Following~\cite{chen2018crowd}, we set the comfort distance to 0.2m; and arrival tolerance, 0.5m. Comfort rate is computed as the percentage of episodes that contain cases of robot getting closer to pedestrians than the comfort distance. Arrival rate is computed as the percentage of episodes that contain cases of robot reaching its goal within arrival tolerance before the cut-off horizon (24 seconds into the future). 

\begin{figure}
    \centering
    \vspace{5pt}
    \includegraphics[scale=0.02]{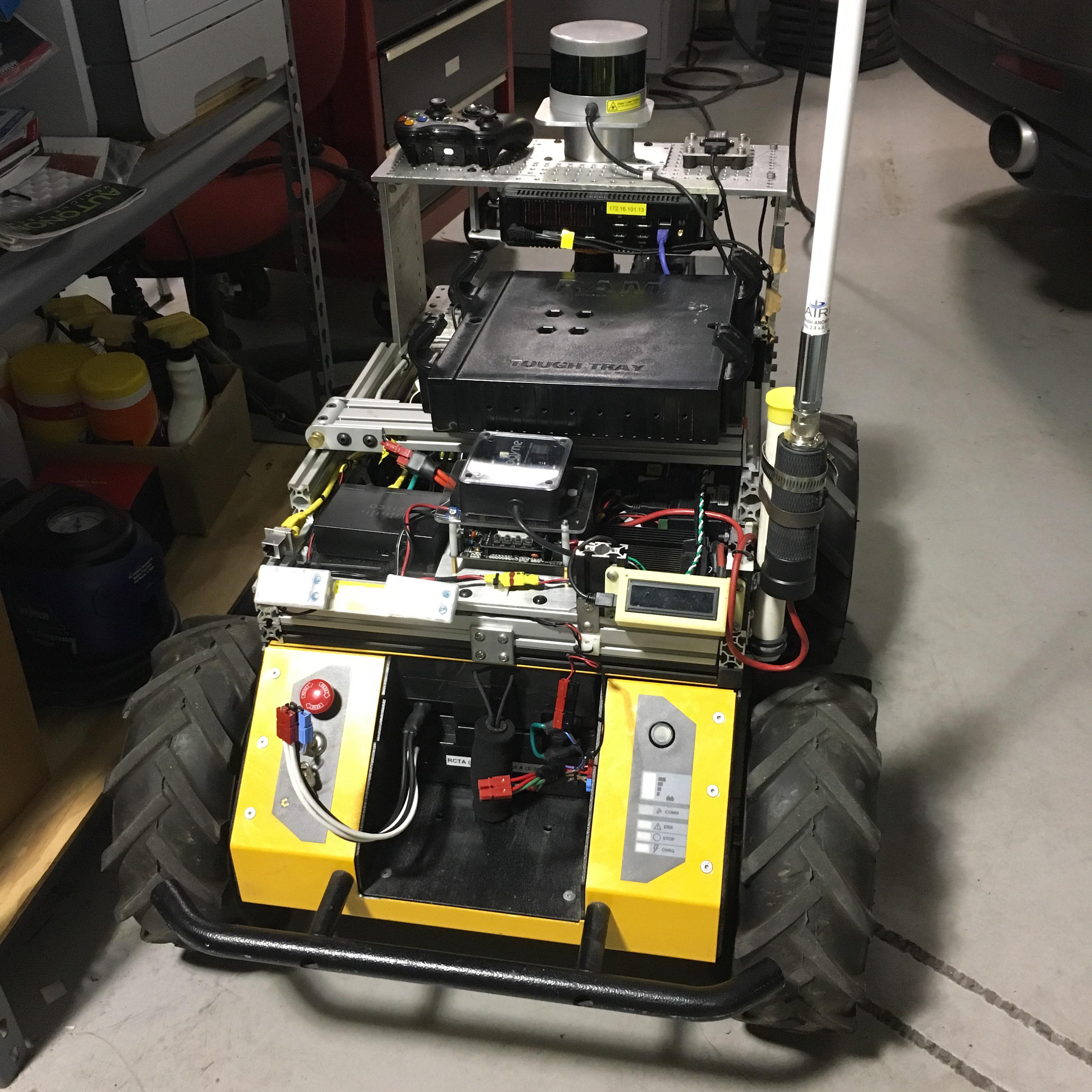}
    \includegraphics[scale=0.2]{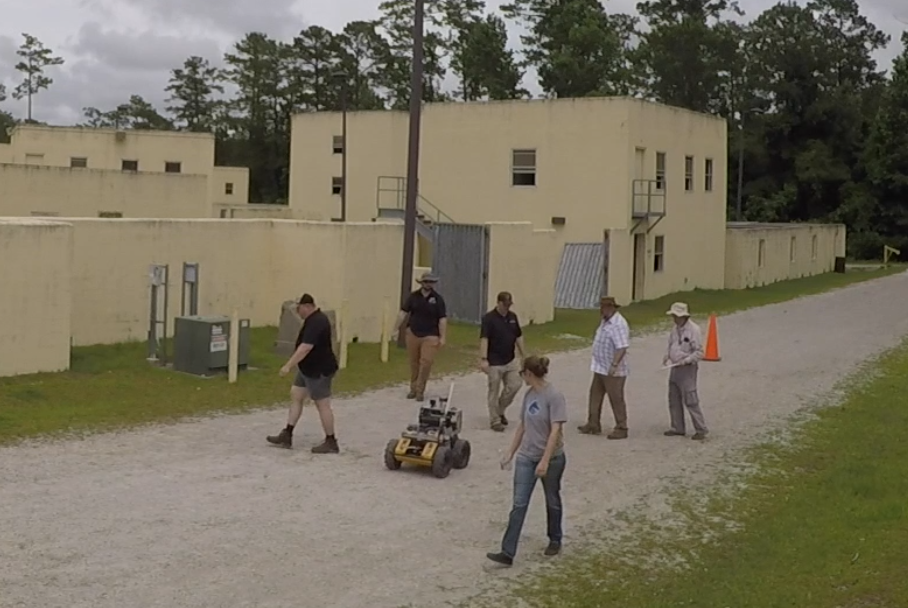}
    \caption{We conduct robotic experiments on a ClearPath Husky robot.}
    \vspace{-20pt}
    \label{fig:husky}
\end{figure}

\textbf{Setup for robot tests:} 
We use a ClearPath Husky robot (Fig. \ref{fig:husky}) with a Nvidia Jetson TX2 development board as our robot platform for field evaluation. 
A lidar-based Kalman filter \cite{welch1995introduction} tracker is applied to identify and keep track of pedestrians in the scene. Beside qualitatively looking at each run of social navigation, we perform two sets of experiments--qualitative measure and A/B testing--to investigate the performance of our approach.

\vspace{-5pt} 
\subsection{Proposed Models} 
 
\noindent \textbf{Goal-Social-LSTM (G)}: A simplified version of NaviGAN trained with the L2 objective, denoted by $\mathcal{L}_2 = \lVert Y_i - \hat{Y}_i \rVert_2$. Goal-Social-LSTM omits the intention-force generator and directly injects the goal state information to the decoder's initial hidden state after the PoolNet layer to predict the state sequence. A diagram of the model is shown in Fig.~\ref{fig:goal_social}.
    
\noindent \textbf{NaviL2 (L2), NaviL2$^\dagger$}: The generators of NaviGAN (block (a) and (b) in Fig. \ref{fig:overview}) trained with both $\mathcal{L}_2$ and final displacement error, \textit{FDE loss}, $\mathcal{L}_{fde}$. Notice that $\mathcal{L}_{fde}$ is applied on the prediction of intention-force generator $\Tilde{Y}_i$. As the final prediction $\hat{Y}_i$ is made by summing $\Tilde{Y}_i$ and $\Vec{F}_{social}$, it is unclear for the model to decide which module should learn the intention and which module should learn the social-awareness. The purpose of $\mathcal{L}_{fde}$ is to eliminate the ambiguity of such composite predictions. The use of the FDE loss guides the intention-force generator toward generating intention force.
    NaviL2 intuitively models $\Vec{F}_{intent}$ and $\Vec{F}_{social}$. We do not include fluctuation noise $\Vec{F}_{fluct}$ when running the social-force generator for NaviL2.
    \begin{equation}
        \mathcal{L}_{fde} = \lVert Y_{T_{end}} - \Tilde{Y}_{T_{end}} \rVert_2
    \end{equation}
    NaviL2$^\dagger$ denotes a model that has only the intention-force generator from a trained NaviL2. The plans generated by NaviL2$^\dagger$ will thus be purely intention-force without considering other agents. The purpose of this intention-force only model is to visualize the target agent's intention for analysis. 

\noindent \textbf{NaviGAN (NG), NaviGAN$^\dagger$, NaviGAN-R (R), NaviGAN-R$^\dagger$}: The complete algorithm with social-aware discriminator trained with  $\mathcal{L}_2$, $\mathcal{L}_{fde}$, and adversarial objective. Analogous to NaviL2$^\dagger$, NaviGAN$^\dagger$ denotes the intention-force generator of NaviGAN. NaviGAN-R is the complete framework jointly trained with $\mathcal{L}_{resist}$. We set 0.5m as $d_{safe}$ for resistance loss.

\vspace{-1pt}
\section{Results}\label{sec:results}
\subsection{Long-term Playback Navigation}\label{exp:long-term}

Since the L2 distance to recorded trajectories does not necessary reflect the social-awareness of a planning algorithm, we propose the use of other statistics to demonstrate the social-awareness. 
Table \ref{tab:s-score} shows the results where S, C$\%$, A$\%$ denote social score, comfort rate, and arrival rate respectively. 
In terms of overall social score from~\cite{chen2018crowd}, humans in the recorded dataset score $0.44$, while all other algorithms score around $0.4$ without a clear winner. 
When we measure the arrival rate and comfort rate separately, NaviGAN shows human-level comfort score while maintaining reasonable arrival rate as well. 

Without $\mathcal{L}_{resist}$, the models focus only on the \textit{naturalness} aspect of socially compliant navigation; therefore, even with high arrival rate, they achieve lower comfort rate. By adding $\mathcal{L}_{resist}$, we optimize for both \textit{naturalness} and \textit{comfort}, and achieve 0.97 comfort rate which is even higher than the recorded human trajectories. By comparing the comfort rate between NaviGAN-R and NaviGAN-R$^\dagger$, we can clearly see the capability of social generator modifying prediction of intention generator to avoid uncomfortable behaviors; however, to enforce \textit{comfort}, NaviGAN-R sometimes can sacrifice the FDE, resulting in a lower arrival rate.

\begin{table}[h!]
    \centering
    \vspace{5pt}
    \caption{Long-term playback navigation result.}
\begin{tabular}{c|cccccccc}
\hline
 & Goal & L2 & L2$^\dagger$ & NG  &NG$^\dagger$& R  &R$^\dagger$& human\\
\hline
S & 0.40 & 0.38 & 0.40 & 0.41 & 0.40 & 0.38 & 0.38 & \textbf{0.44}\\
C$\%$& 0.81 & 0.81 & 0.82 & 0.82 & 0.80 & \textbf{0.97} & 0.85 & 0.96 \\
A$\%$&0.91 & 0.88 & 0.92 & 0.97 & 0.92 & 0.85 & 0.88 & \textbf{1.00}\\
\hline
\end{tabular}
\vspace{-15pt}

    \label{tab:s-score}
\end{table}

\begin{figure*}[htbp]
    \centering%
    \makebox[\textwidth][c]{\includegraphics[width=0.8\textwidth]{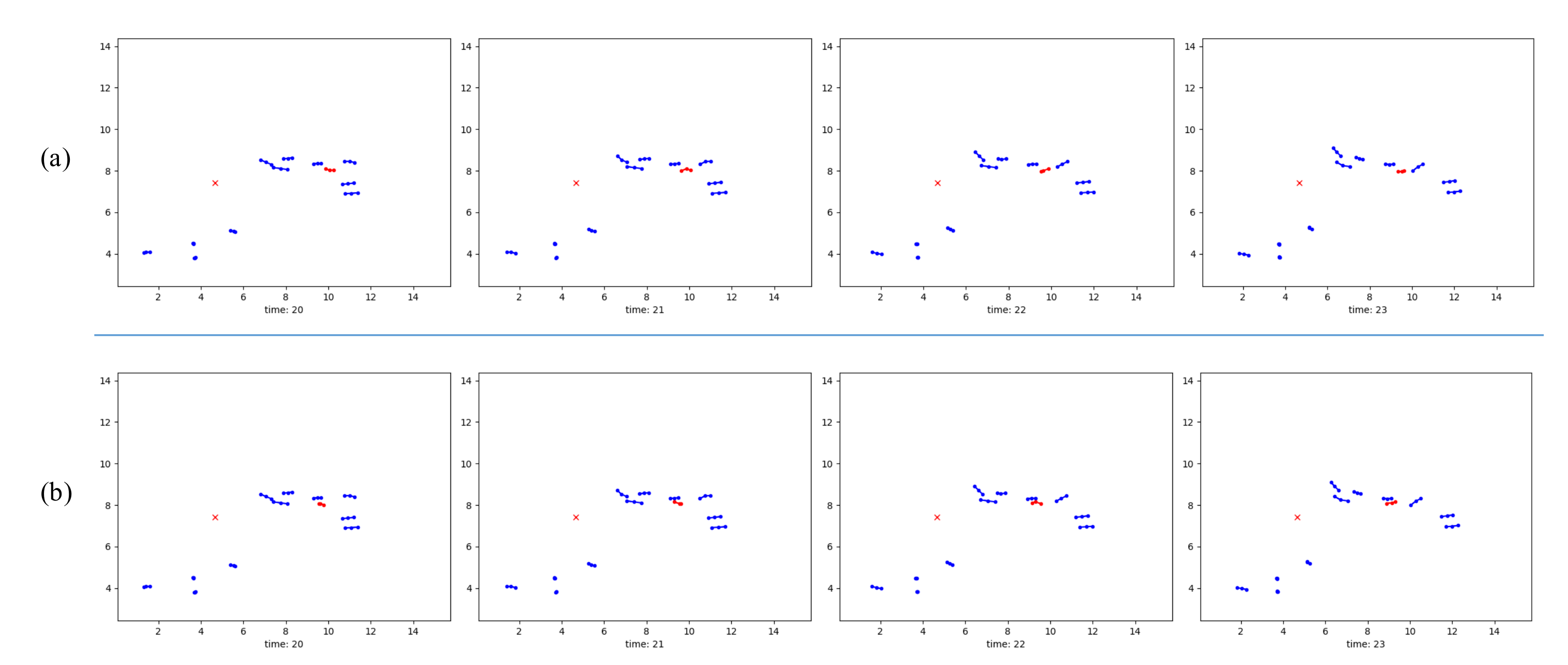}}%
    \vspace{-10pt}
    \caption{An example where the social-force generator affects the sequence predicted by the intention-force generator to enforce social norm such as avoiding to get uncomfortably close to others, while still pursuing a goal (marked as a red cross). We visualize the state sequence at time $t$ by showing its two previous states along with the current state in a dotted-line. Red dots are the footprints of our target agent, and blue dots are the footprints of other agents. Subfigure (a) visualizes the path planned by NaviGAN, and (b) visualizes the path planned by NaviGAN$^\dagger$ (intention-force only).}%
    \vspace{-10pt}
    \label{fig:maintain_distance}%
\end{figure*}
%
        
        

\textbf{Interpretability:} For better understanding of the relationship between the intention-force generator and the social-force generator, we visualize an example of episode rolled out by NaviGAN and NaviGAN$^\dagger$ in Fig. \ref{fig:maintain_distance}. 
We can clearly see how the social-force generator adjusts the sequence predicted by the intention-force generator to enforce social-compliance, \exempli maintain comfortable distance to others, while still moving toward a goal state.

\subsection{Robot Experiments}\label{sec:robot-exp}

\subsubsection{Qualitative Measurement}
As a qualitative measurement, we systematically design scenarios to interact with the robot running our socially compliant navigation module. We score each run with one of the three types of score: \textbf{1}--Collision with other pedestrian happens during run. The behavior is obviously not socially compliant. \textbf{2}--No collision happens. Robot demonstrates not natural behavior and only mild social compliance is observed. \textbf{3}--The robot complies with social rules in a natural and comfort manner.

The experimental design was carried out on JMP 13 software by SAS Inc. The experimental plan was
structured as a Main effects design, completely randomized D-optimal design \cite{de1995d} in 27 runs\footnote{\vspace{-5pt} The robot test video is available at: \href{https://youtu.be/61blDymjCpw}{https://youtu.be/61blDymjCpw}.}. The inputs for the design were: Number of Pedestrians (2, 4, 6), Walking Direction (Parallel to the path of the
oncoming robot, Parallel and Crossing the path, Crossing), and Crossing Distance (meters) that pedestrians
crossed in front of the oncoming bit (1, 2, 3) in meters. Pedestrians was entered as ``discrete numeric'',
Direction as ``categorical'', and crossing distance as ``numeric.''  In ``Parallel $\&$ Crossing,'' some pedestrians
crossed the path of the robot (at approximately 1, 2, or 3 meters distance) while others (at outer edge)
continued on a parallel path. 

A goal was always set to be on the right of the orange cone. In all runs, the robot never collided with pedestrians--i.e., score 1 never happened--and it always reached the goal within 0.5 meter tolerance. Among all the trials, the algorithm achieved 86.7$\%$ of score 3, and 0$\%$ of score 1. In most of the failure cases (cases that score 2, e.g., run $\#$5 and $\#$9), unnatural behavior happened only after interaction ( i.e., all pedestrians have passed) and probably was due to the platform's mobility limitations, e.g., the robot was spinning in-place before carrying on toward a goal. Experiment $\#$17 is a particularly interesting case where the robot actually rolled back to avoid collision before the space gets cleared off and it continued to navigate. This kind of timid behavior was marked as unnatural by a human evaluator. 



\subsubsection{A/B Testing}

\begin{table}[h!]
    \centering
    \caption{A/B testing result.}
    \vspace{-3pt}
    \begin{tabular}{r|cc}
    \hline
         & SBPL & NaviGAN-R \\
         \hline
        $\#$peds $\geq$ 2 & 67$\%$ & \textbf{67$\%$}\\
        $\#$peds $\geq$ 4 & 64$\%$ & \textbf{75$\%$}\\
        $\#$peds $\geq$ 6 & 58$\%$ & \textbf{75$\%$}\\
    \hline
    \end{tabular}
    \label{tab:ABtest}
    \vspace{-15pt}
\end{table}

To demonstrate the ability of our approach to optimize both \textit{naturalness} and \textit{comfort}, we compare our approach against pure collision avoidance algorithm that only optimizes the \textit{comfort} aspect of socially compliant navigation. We design an A/B test experiment where we either use our approach or pure collision avoidance as controller to interact with subjects and ask them to identify whether they were interacting with socially compliant controller or not.

The test plan was structured as a Main-and-two-factor-interaction plus quadratic, completely randomized D-optimal design in 18 runs. The input for the design was unchanged except that Crossing Distance was set to 2m  and that Controller (A, B) was used as additional input. The execution of the experiment was ``blinded'' in that the participants did not know which controller was active during each run.

For the collision-avoidance controller, we use the popular search based planning library (SBPL)~\cite{SBPL}. The results are summarized in Table~\ref{tab:ABtest}. Controller A is a collision avoidance algorithm; and controller B, our proposed model. Notice that in all runs, either with collision avoidance or our approach as controller, the robot never collided with a pedestrian and it always reached the goal within 0.5m tolerance. 
The benefit of social compliance is not obvious when there are only a few pedestrians, but the results on an increased number of pedestrians support that our approach of socially compliant navigation is strongly preferred to the baseline. 

\vspace{-5pt}
\section{Conclusion}
\vspace{-3pt}
We present NaviGAN, a data-driven, deep generative approach for social navigation that considers both \textit{comfort} and \textit{naturalness} of social compliance. The proposed modeling of the intention, social, and fluctuation forces as specially designed LSTMs makes NaviGAN highly interpretable as an agent's intention can be conveniently visualized. 
Although direct performance comparison is not possible due to different experimental settings, in comparison to reinforcement learning based methods that learns a policy to maximize the expectation of a hand-crafted reward function, NaviGAN takes a purely data-driven approach to discover the latent social interaction model encoded in human navigation data through adversarial training.
We demonstrate the advantages of NaviGAN through an extensive set of experiments. The results highlight the benefits of resistance force in particular as well as the interpretable behaviors. 
A variation of NaviGAN has been successfully deployed on multiple robot platforms, \exempli a wheelchair robot~\cite{yao}; we plan to study the impact of the physical appearance of robots in real-world scenarios.

\end{document}